\documentclass[10pt, a4paper]{article}

\usepackage[final]{lrec2026} 

\usepackage[utf8]{inputenc}        
\usepackage[T1]{fontenc}           

\usepackage{times}                 
\usepackage{latexsym}              
\usepackage{inconsolata}           

\usepackage{amsmath, amssymb}      
\usepackage{amsfonts}              
\usepackage{bm}                    

\usepackage{booktabs}              
\usepackage{multirow}              
\usepackage{array}                 
\usepackage{siunitx}               

\usepackage{graphicx}              
\usepackage{caption}               
\usepackage{float}
\usepackage{subcaption}

\usepackage{enumitem}              

\usepackage{nicefrac}              
\usepackage{microtype}             

\usepackage{hyperref}               
\usepackage{url}                    

\title{Orthographic Constraint Satisfaction and Human Difficulty Alignment in Large Language Models}

\name{Bryan E. Tuck, Rakesh M. Verma} 

\address{University of Houston \\
         Houston, Texas \\
         betuck@uh.edu, rmverma2@central.uh.edu\\}

\abstract{
Large language models must satisfy hard orthographic constraints during controlled text generation, yet systematic cross-family evaluation remains limited. We evaluate 39 configurations spanning three model families (Qwen3, Claude Haiku 4.5, GPT-5-mini) on 58 word puzzles requiring character-level constraint satisfaction. Cross-family differences produce substantially larger performance gaps (2.0--2.2$\times$, $F_1$ =0.761 vs.\ 0.343) than parameter scaling within families (83\% gain from 4B to 32B scaling), and a partial-correlation analysis rules out tokenizer design as a confound for within-family scaling. Thinking budget sensitivity proves heterogeneous: high-capacity models show strong returns (+0.102 to +0.136 $F_1$), while mid-sized variants saturate or degrade, showing inconsistent compute benefits. Using difficulty ratings from 10,000 human solvers per puzzle, we establish modest but consistent calibration ($\rho$ = 0.28--0.42) across all families, yet identify systematic failures on common words with unusual orthography (``data'', ``loll'', ``acai'': 83--91\% human success, 94--98\% model miss rate). These failures point to over-reliance on distributional plausibility that penalizes orthographically atypical but constraint-valid patterns.
 \\ \newline  }

\begin{document}

\maketitleabstract

\section{Introduction}

How do large language models satisfy hard orthographic constraints during text generation? Consider a model asked to generate English words using only the letters \{A, G, I, L, N, O, W\}, where every word must contain W and have at least four letters. This task requires satisfying discrete character-level rules, a fundamental challenge distinct from the semantic pattern matching that dominates language model training~\cite{Pulvermller2010BrainLanguageRW}. Unlike classification tasks with predefined outputs, constrained generation demands navigation of large combinatorial spaces while maintaining strict constraint adherence~\cite{garbacea2025why, adv_nlp_survey, yin-neubig-2022-interpreting}. Characterizing how neural networks handle symbolic constraints informs applications requiring structured output (educational word games, controlled content generation, formal language tasks) while probing whether current architectures represent linguistic structure beyond distributional patterns.\footnote{Code: \url{https://github.com/ReDASers/spelling-bee-llm-eval}. \\Spelling-Bee Dataset: \url{https://huggingface.co/datasets/redasers/spelling-bee-human-difficulty}}

Despite extensive research on language model generation capabilities, few studies examine how models handle hard orthographic constraints in open-vocabulary settings. Prior work on constrained text generation has largely focused on semantic or lexical constraints~\cite{lu-etal-2022-neurologic}, while recent benchmarks suggest models still struggle with explicit character-level rules~\cite{edman-etal-2024-cute}. Little prior work has systematically evaluated how model scale, computational budgets, and reasoning modes interact to satisfy character-level constraints with human-calibrated difficulty data.

A core tension in constrained generation is the conflict between distributional plausibility and structural validity~\cite{dziri2023faithfatelimitstransformers, andreas-2022-language}. Models trained on distributional patterns may assign low generation probabilities to constraint-valid solutions with atypical orthographic patterns~\cite{liu2023transformerslearnshortcutsautomata}: a word may be frequent in training data yet appear implausible under specific letter restrictions, or satisfy all constraints yet trigger low generation probabilities due to unusual letter combinations.

We address this gap using word puzzles from the New York Times Spelling Bee, where models generate English words using only specified letter sets with mandatory letter inclusions and minimum length requirements. This task provides three methodological advantages: (1) it isolates constraint-handling from semantic reasoning, enabling focused study of orthographic knowledge; (2) difficulty is calibrated for human solvers, providing ecological validity for model-human comparison; and (3) solver data from 10,000 solvers per puzzle enables direct calibration without proxy measures.

We conduct an evaluation across three model families spanning 39 configurations from Qwen3, Claude Haiku 4.5, and GPT-5-mini, testing direct generation and reasoning-guided modes with varying computational budgets. Across 58 puzzles, we evaluate 2,262 experiments (39 configurations $\times$ 58 puzzles) in zero-shot settings to isolate intrinsic constraint-handling capabilities.

Our contributions are as follows:

\begin{enumerate}
\item \textbf{Cross-family performance characterization} (Section~\ref{sec:arch_scale}): Cross-family differences substantially exceed within-family scaling gains, with the gap manifesting primarily through recall rather than precision. While the drivers (architecture, training data, or optimization) cannot be fully isolated, a partial-correlation analysis (Appendix~\ref{sec:tokenization_check}) rules out tokenizer design as a within-family confound.

\item \textbf{Heterogeneous budget sensitivity} (Section~\ref{sec:budget_sensitivity}): Thinking budget effects vary dramatically across models: high-capacity variants show strong returns, mid-size models degrade monotonically with increased allocation, and the mixture-of-experts 30B variant requires a substantial computational budget to compensate for its limited effective capacity.

\item \textbf{Human difficulty alignment with failure analysis} (Sections~\ref{sec:human_alignment},~\ref{sec:length_patterns}): Using 10,000 solvers per puzzle, we establish modest but consistent model-human calibration across families, yet identify systematic failures on common words with unusual orthography that point to over-reliance on distributional plausibility.
\end{enumerate}

\section{Related Work}

\textbf{Orthographic Knowledge in Language Models.}
Language models encode orthographic structure beyond token frequencies~\cite{itzhak2022spellingbee}, yet limitations persist: multilingual LLMs over-weight orthographic similarity when processing interlingual homographs~\cite{tanwar2025orthosemantics}, show systematic gaps in grapheme-to-phoneme mapping~\cite{suvarna2024phonologybench}, and exhibit brittleness to character-level perturbations from subword tokenization~\cite{chai2024subword}. While character-level~\cite{bunzeck-etal-2024-graphemes} and byte-level approaches~\cite{dang-etal-2025-tokenization} avoid tokenization artifacts, they remain less common than subword-based architectures. These findings reveal that orthographic knowledge in LLMs remains fragmented and inconsistent, motivating systematic evaluation under explicit constraint satisfaction rather than recognition tasks alone.

\textbf{Constrained Generation Methods.}
Existing approaches primarily employ decoding-time guarantees through Grid Beam Search~\cite{hokamp-liu-2017-lexically}, DOMINO~\cite{beurerkellner2024domino}, and grammar-constrained generation~\cite{grammarconstraint2025,park2025flexibleefficientgrammarconstraineddecoding}, which guarantee constraint satisfaction by restricting the token space during decoding. We test whether instruction-tuned models can satisfy constraints through prompting alone, without decoding-time enforcement.

\textbf{Evaluation with Word Puzzles and Human Difficulty.}
Word puzzle tasks provide structured testbeds for constraint satisfaction~\cite{giadikiaroglou-etal-2024-puzzle}. Recent work on crossword solving~\cite{Saha_2025} demonstrates LLM capabilities on character-constrained puzzles requiring length adherence and character overlaps. Work on lexical complexity prediction~\cite{kelious-etal-2024-complex, nohejl-etal-2025-beyond} shows models can assess word difficulty, but evaluations typically lack human difficulty baselines for generative tasks. Existing lexical complexity evaluations use frequency-based features with small annotator pools~\cite{kelious-etal-2024-complex} or word frequency norms as proxies for human familiarity~\cite{nohejl-etal-2025-beyond}. We instead incorporate behavioral data from 10,000 solvers per puzzle to enable direct model-human calibration on constraint satisfaction performance.

\section{Experimental Setup}

\subsection{Task Definition}

Models generate English words satisfying explicit orthographic constraints. Each puzzle specifies seven unique letters, one designated as mandatory. Valid outputs must satisfy three constraints: minimum four letters, exclusive use of the seven available letters (repetition permitted), and mandatory inclusion of the designated center letter. For example, given \{A, G, I, L, N, O, W\} with W as mandatory, ``wagon'' is valid (uses only available letters, includes W, length $\geq$ 4) while ``along'' is invalid (missing W) and ``awning'' is valid (uses A, W, N, I, N, G from available set, with N repeated).

Words using all seven letters are designated ``pangrams.'' Puzzles are drawn from a professionally curated collection to ensure vocabulary diversity and appropriate difficulty for human solvers.

\subsection{Dataset}

We evaluate models on 58 consecutive daily puzzles from the New York Times Spelling Bee (June~2--July~29, 2025), containing 2,710 total word instances spanning 2,007 unique words across curated, human-verified solution sets.Table~\ref{tab:dataset_stats} presents comprehensive dataset statistics, including dataset scale, solution set sizes per puzzle, word length distributions, and pangram counts.

\begin{table}[t]
\centering
\small
\resizebox{\linewidth}{!}{
\begin{tabular}{l@{\hspace{0.5em}}c@{\hspace{0.5em}}c}
\toprule
\textbf{Property} & \textbf{Value/Mean} & \textbf{Range} \\
\midrule
Words per puzzle & $46.7 \pm 12.1$ & 22--72 \\
Pangrams per puzzle & $1.62 \pm 0.83$ & 1--4 \\
Word length (letters) & $5.52 \pm 1.59$ & 4--13 \\
\midrule
4-letter words & 940 & (34.7\%) \\
5-letter words & 615 & (22.7\%) \\
6+ letter words & 1,155 & (42.6\%) \\
\bottomrule
\end{tabular}
}
\caption{Puzzle and vocabulary characteristics. Solution sets average 47 words with substantial variability (22--72), and longer words (6+ letters) comprise 43\% of targets.}
\label{tab:dataset_stats}
\end{table}

\subsection{Human Difficulty Data}

We augment each puzzle with aggregate performance statistics from the NYT Spelling Bee platform, where 10,000 solvers attempt each daily puzzle. For each word, we obtain solver success rates (range: 3--97\%), providing ground-truth human difficulty estimates calibrated to actual solver performance rather than proxy measures. The platform provides optional solver aids, including a grid of remaining words organized by starting letter and length, and a two-letter frequency list. Reported success rates therefore reflect potentially hint-assisted performance, which may inflate success rates relative to unaided solving.

\subsection{Models and Configurations}

We evaluate three model families to examine constrained generation across scales, architectures, and reasoning mechanisms. The selection enables controlled comparison: Qwen3 provides open-source models spanning parameter scales with both dense and mixture-of-experts architectures, while Claude Haiku 4.5 and GPT-5-mini represent proprietary systems with different reasoning implementations, allowing us to compare cross-family differences against within-family scaling effects.

\textbf{Qwen3 Family.} 
Five open-source Qwen3 models~\cite{yang2025qwen3technicalreport} span 4B to 32B parameters: four dense transformers (4B, 8B, 14B, 32B) and one Mixture-of-Experts model (30B-A3B with 3B active parameters). This family enables within-architecture scaling analysis while the MoE variant tests whether sparse expert routing affects constraint satisfaction. We test each model in direct generation mode and thinking mode, both at three token budgets (4K, 8K, 16K), yielding 30 configurations (5 models $\times$ 6: 3 direct + 3 thinking). Thinking mode generates reasoning traces before outputs; direct mode produces responses immediately. Varying the output budget in direct mode controls maximum generation length, which affects the number of words produced.

\textbf{Claude Haiku 4.5.}
We evaluate Claude Haiku 4.5 \citep{Anthropic_Claude4_SystemCard_2025} in direct generation mode at three budgets (4K, 8K, 16K tokens) and extended thinking at three budgets (4K, 8K, 16K tokens), yielding 6 configurations.

\textbf{GPT-5-mini.}
We test GPT-5-mini~\cite{OpenAI_GPT5_SystemCard_2025} at three reasoning effort levels (low, medium, high), yielding 3 configurations. All GPT-5-mini configurations use reasoning; no direct generation baseline is available. 

\begin{figure}[t]
\centering
\includegraphics[width=0.45\textwidth]{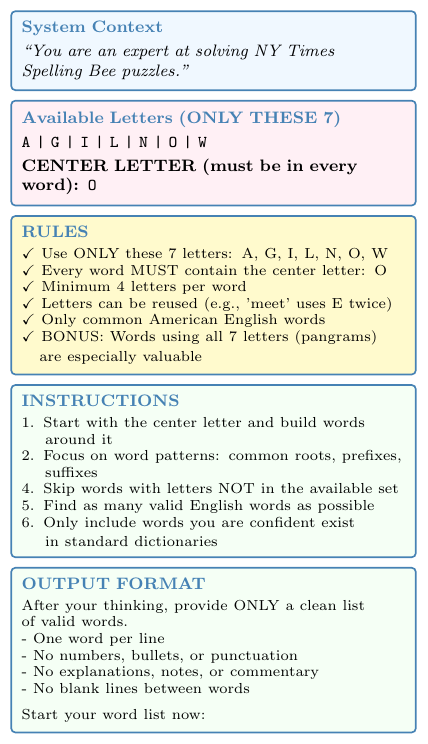}
\caption{Zero-shot prompt structure. The prompt specifies the seven available letters, marks the mandatory center letter, and enumerates all constraints explicitly. Models receive identical specifications without solution counts, isolating intrinsic constraint-handling from memorization or calibration to expected output lengths.}
\label{fig:prompt_structure}
\end{figure}
 
\subsection{Prompt Design}

Figure~\ref{fig:prompt_structure} illustrates the zero-shot prompt structure. All models receive identical prompts specifying the seven available letters, mandatory center letter, and three constraints. Prompts use a target-free formulation (``Find as many valid English words as possible'') without revealing solution counts, and request one word per line as output.

\subsection{Implementation Details}

Qwen3 models run on vLLM~\cite{kwon2023efficient} (NVIDIA A100 GPUs) with Qwen-recommended sampling parameters: temperature 0.6, top-p 0.95, top-k 20, and repetition penalty 1.2 for thinking mode; temperature 0.7, top-p 0.8, top-k 20, and repetition penalty 1.2 for direct mode. Thinking budgets are enforced through two-stage generation: stage one generates up to the budget limit, and if the thinking trace is not complete, an early-stop prompt is injected to trigger the answer. Claude-Haiku uses the Anthropic API with native extended thinking support, where \texttt{budget\_tokens} directly caps thinking length and temperature is fixed at 1.0. GPT-5-mini uses the OpenAI Responses API with three qualitative reasoning effort levels (low, medium, high); we map these to approximate budget equivalents (4K, 8K, 16K) for cross-model comparison. Claude-Haiku and GPT-5-mini use API-default sampling parameters.

\section{Evaluation Metrics}

\subsection{Standard Metrics}

We evaluate generation quality using precision, recall, and F1 score. For puzzle $i$, let $G_i$ denote generated words and $S_i$ denote the verified solution set. Precision is $P = |G_i \cap S_i| / |G_i|$, recall is $R = |G_i \cap S_i| / |S_i|$, and F1 score is $F_1 = 2PR / (P + R)$.

\subsection{Human Difficulty Alignment}

Using solver data from 10,000 solvers per puzzle, we define human difficulty for each word-puzzle occurrence as $d_h(w, p) = 1 - (\text{user\_success\_count}(w, p) / 10{,}000)$, ranging from 0.03 to 0.97 across 2,710 word-puzzle instances. Words appearing in multiple puzzles receive independent difficulty scores per occurrence, as the surrounding letter set affects solver success rates.

We compute two alignment metrics: (1) calibration strength, Spearman rank correlation between human difficulty and model difficulty (fraction of configurations in which the model missed each word), and (2) quartile stratification, model recall across human difficulty quartiles (Q1: easiest 25\%, Q4: hardest 25\%).

\subsection{Length-Stratified Analysis}

We stratify recall by word length (4-letter, 5-letter, 6-letter, 7+ letters) to identify length-dependent failure patterns. Recall is computed as the fraction of words in each length range successfully generated, aggregated across all puzzles. Section~\ref{sec:length_patterns} shows that models degrade catastrophically with word length (up to 71$\times$ for small models) while humans decline only 1.3$\times$.

\section{Results}

\subsection{Cross-Family and Scale Effects on Constraint Satisfaction}
\label{sec:arch_scale}

\begin{figure}[t]
\centering
\includegraphics[width=0.92\columnwidth]{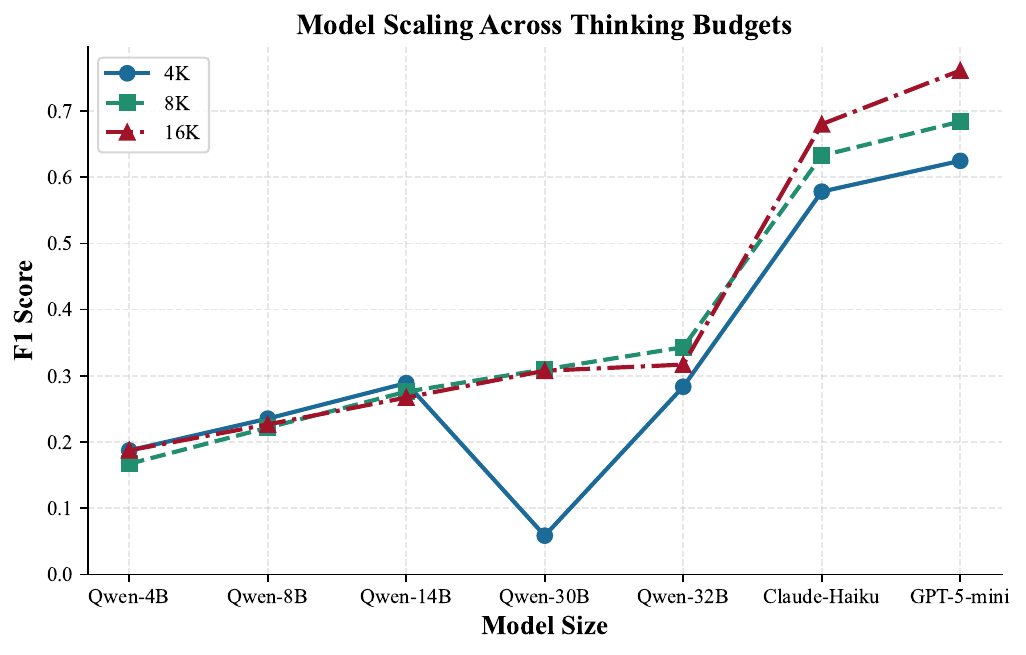}
\caption{Cross-family performance comparison across thinking budgets. Proprietary models achieve 2.0--2.2$\times$ higher F1 than the largest open-source model, with the gap driven primarily by recall (68\% vs.\ 23\%) rather than precision. Budget sensitivity varies across families.}
\label{fig:scaling_budget}
\end{figure}

Figure~\ref{fig:scaling_budget} reveals a performance hierarchy across model families. Proprietary models achieve F1 scores 2.0--2.2$\times$ higher than the largest open-source configuration tested. Within the Qwen family, scaling from 4B to 32B yields an 83\% F1 improvement (0.187 to 0.343), far less than the cross-family gap. Constraint satisfaction depends on factors beyond parameter scaling alone, though the specific drivers (architectural innovations, training data composition, tokenizer design, or optimization procedures) cannot be disentangled without access to proprietary details.

\paragraph{Precision-Recall Decomposition of Performance Differences}

Across all 39 configurations, no model produced a single word using letters outside the specified set in their final respective output (Appendix~\ref{sec:pangram_fp}). This perfect exclusion compliance, verified programmatically by checking each generated word against the puzzle letter set, reveals an asymmetry: models flawlessly verify ``is this letter allowed?'' yet fail to retrieve valid words from the same letter set. The constraint satisfaction bottleneck lies in generation under constraints rather than constraint checking itself.

The performance gap manifests primarily through recall rather than precision (Table~\ref{tab:cross_family}). Proprietary models discover far more of the valid solution space, while precision differences remain modest ($\approx$9 percentage points). The gap widens on pangrams, which require all seven letters: GPT-5-mini achieves 77.6\% pangram recall versus Qwen-32B's 15.8\%, a 4.9$\times$ ratio that exceeds the overall F1 gap (Appendix~\ref{sec:pangram_fp}). The quality gap extends beyond recall: 64.6\% of GPT-5-mini's false positives are valid English words excluded by NYT editorial curation, compared to 13.3\% for Qwen, indicating that raw precision understates the quality gap between families.

\begin{table}[t]
\centering
\small
\resizebox{\linewidth}{!}{
\begin{tabular}{l@{\hspace{0.4em}}r@{\hspace{0.4em}}r@{\hspace{0.4em}}r@{\hspace{0.4em}}r@{\hspace{0.4em}}r}
\toprule
\textbf{Model} & \textbf{Budget} & \textbf{P} & \textbf{R} & \textbf{F1} & \textbf{Vol} \\
\midrule
GPT-5-mini    & 16K & \textbf{0.888} & \textbf{0.680} & \textbf{0.761}{\scriptsize$\pm$0.073} & \textbf{34.4} \\
Claude-Haiku  & 16K & 0.851 & 0.574 & 0.680{\scriptsize$\pm$0.108} & 30.4 \\
Claude-Haiku  & 8K  & 0.842 & 0.515 & 0.633{\scriptsize$\pm$0.104} & 27.9 \\
GPT-5-mini    & 8K  & 0.839 & 0.590 & 0.684{\scriptsize$\pm$0.125} & 31.3 \\
Qwen-32B      & 8K  & 0.797 & 0.233 & 0.343{\scriptsize$\pm$0.154} & 12.7 \\
Qwen-30B      & 8K  & 0.786 & 0.201 & 0.310{\scriptsize$\pm$0.134} & 11.2 \\
Qwen-14B      & 4K  & 0.768 & 0.184 & 0.289{\scriptsize$\pm$0.116} & 10.6 \\
\bottomrule
\end{tabular}
}
\caption{Highest-performing configurations across model families, ranked by F1. F1 values include standard deviation across 58 puzzles. Vol is the average number of valid words generated per puzzle.}
\label{tab:cross_family}
\end{table}

\paragraph{Benefits of Enabling Thinking Mode}

Enabling thinking improves performance across all tested models, though the mechanism differs from conventional expectations (Table~\ref{tab:thinking_effects}). Improvements manifest primarily through precision rather than recall, particularly at smaller scales (the gap narrows for larger models), the opposite of what expanding solution search would predict. Thinking mode reduces false positives more than it expands solution coverage. This verification-dominant pattern holds across the Qwen family and extends to Claude-Haiku, indicating that extended reasoning strengthens constraint checking rather than broadening lexical search. This precision-first mechanism predicts that budget sensitivity (Section~\ref{sec:budget_sensitivity}) should track verification load: models generating more false positives at baseline stand to gain most from additional reasoning tokens.

\subsection{Model-Dependent Budget Sensitivity}
\label{sec:budget_sensitivity}

\begin{figure}[t]
\centering
\includegraphics[width=0.95\columnwidth]{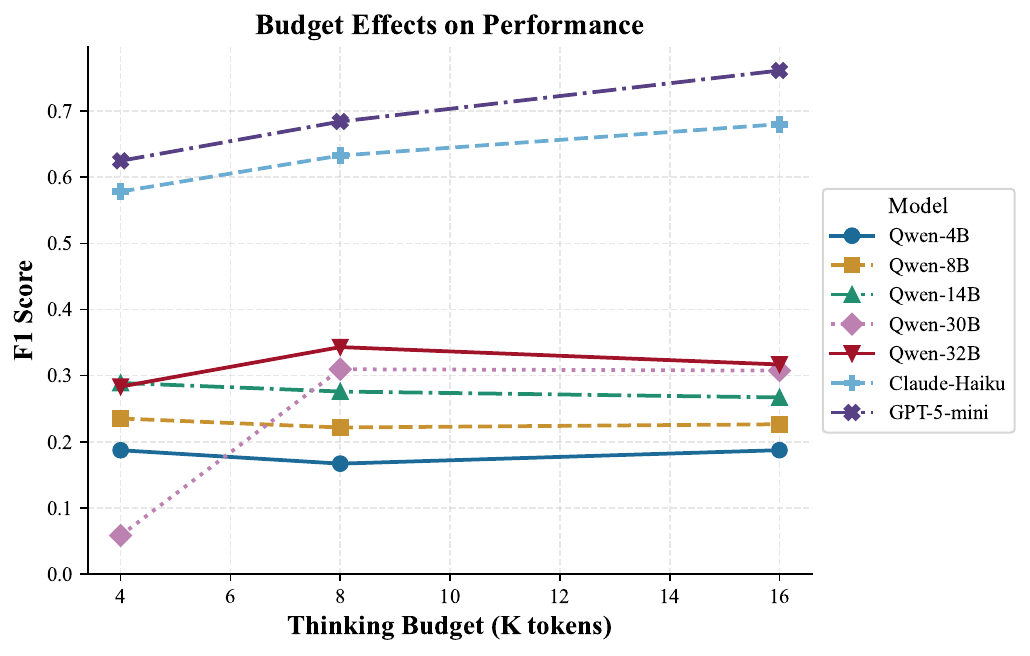}
\caption{Heterogeneous budget sensitivity across model sizes. Smaller models remain flat with additional budget, while the 14B variant paradoxically degrades. The MoE 30B variant (3B active parameters) shows the strongest budget dependence ($\Delta$F1=+0.249). Proprietary systems show consistent improvements.}
\label{fig:budget_effects}
\end{figure}

Figure~\ref{fig:budget_effects} illustrates heterogeneous budget trajectories across the Qwen3 family and proprietary models, revealing four distinct behavioral classes. Optimal thinking budgets diverge across models, with some showing zero or negative returns from additional tokens.

\paragraph{Responsive High-Capacity Models: Dense and MoE Dynamics}

Within the Qwen3 family, budget responsiveness (F1 improvement with increased thinking budget) appears only in the high-capacity variants (30B and 32B), though their architectures differ fundamentally: the 30B employs mixture-of-experts (MoE) while the 32B uses a dense architecture.

\begin{table}[t]
\centering
\small
\resizebox{\linewidth}{!}{
\begin{tabular}{l@{\hspace{0.4em}}r@{\hspace{0.4em}}r@{\hspace{0.4em}}r@{\hspace{0.4em}}r}
\toprule
\textbf{Model} & \textbf{$\Delta$P} & \textbf{$\Delta$R} & \textbf{$\Delta$F1} & \textbf{Rel. (\%)} \\
\midrule
Qwen-4B   & \textbf{+0.454} & +0.080 & +0.134 & +291 \\
Qwen-8B   & +0.302 & +0.109 & +0.172 & +310 \\
Qwen-14B  & +0.301 & +0.140 & +0.214 & \textbf{+335} \\
Qwen-30B  & +0.155 & +0.087 & +0.130 & +137 \\
Qwen-32B  & +0.187 & +0.141 & +0.197 & +169 \\
\midrule
Claude-Haiku & +0.414 & \textbf{+0.208} & \textbf{+0.290} & +79 \\
\bottomrule
\end{tabular}
}
\caption{Thinking mode effects averaged across matched budgets ($\Delta$ = Thinking ON $-$ Thinking OFF). GPT-5-mini is excluded because all tested configurations use reasoning (no OFF baseline).}
\label{tab:thinking_effects}
\end{table}

The MoE variant exhibits particularly dramatic budget dependence, underperforming other family members at minimal allocation yet improving from $F_1$ =0.058 to $F_1$ =0.310 in the 4K-to-8K range before plateauing ($\Delta$F1=+0.249). The dense 32B, by contrast, shows only modest sensitivity ($\Delta$F1=+0.033 across the same range). This asymmetry is consistent with active parameter count as the driving factor: the 30B-A3B activates only 3B parameters per token, the smallest effective capacity of any model tested, requiring more reasoning steps to compensate.

Both patterns carry a practical deployment consequence: large models with insufficient budget may yield worse performance than smaller models at the same allocation, inverting typical parameter-scaling assumptions.

\begin{figure*}[t]
\centering
\includegraphics[width=0.96\textwidth]{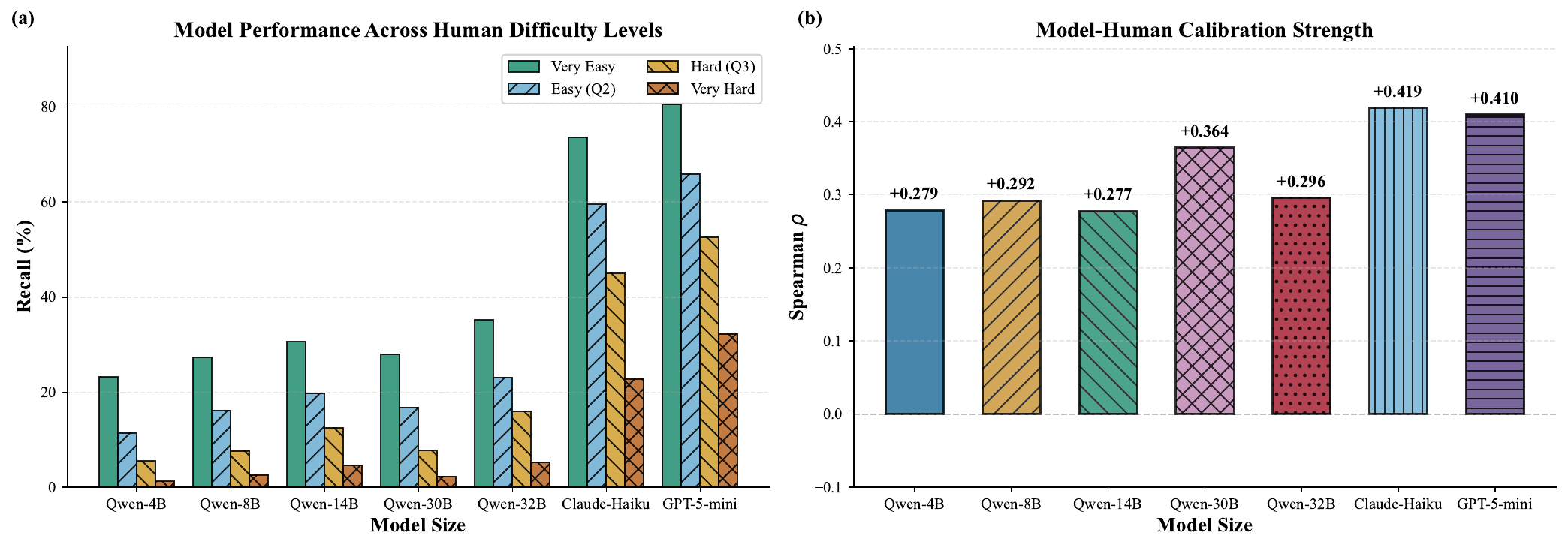}
\caption{Model-human difficulty calibration using 10,000 solvers per puzzle. Left: Performance gradients from easy to hard words vary by model capacity (19$\times$ drop for Qwen-4B vs.\ 2.5$\times$ for GPT-5-mini). Right: Calibration strength (Spearman $\rho$) at each model's F1-maximizing budget shows modest alignment ($\rho$ = 0.28--0.42, all $p < 0.001$), with proprietary models achieving higher correlations.}
\label{fig:human_calibration}
\end{figure*}

\paragraph{Budget-Insensitive Small Models}

The smallest variants (4B and 8B) exhibit budget insensitivity, maintaining flat performance across the tested range. Without thinking mode, Qwen-4B produces zero valid words on 51\% of puzzles, a complete generation collapse absent from all proprietary configurations (Appendix~\ref{sec:strategy_profiles}). These models lack the representational capacity to decompose constraint satisfaction into profitable sub-steps, and additional budget cannot compensate.

\paragraph{Paradoxical Degradation in Mid-Sized Models}

Unlike the small models that gain nothing from budget and the large models that benefit, the 14B variant actively degrades as reasoning budget increases. One explanation connects to the distributional plausibility bottleneck identified in Section~\ref{sec:length_patterns}: for a mid-capacity model, extended thinking may amplify distributional generation without proportionally improving constraint verification, producing more candidates that appear plausible but violate constraints. This interpretation aligns with the Qwen volume-increasing pattern (Appendix~\ref{sec:strategy_profiles}), where more budget generates more candidates rather than better-verified ones.

\paragraph{Proprietary Model Budget Efficiency}

Proprietary models (Claude-Haiku, GPT-5-mini) show consistent positive returns across the full budget range without saturation or degradation. The appendix generation strategy analysis (Appendix~\ref{sec:strategy_profiles}) reveals a mechanistic explanation: Claude reduces output volume by 19\% under thinking while doubling F1, a volume-decreasing pattern where additional tokens refine the candidate set. Qwen models, by contrast, increase output volume under thinking. The volume-decreasing pattern scales better because additional reasoning tokens appear to eliminate false positives, whereas volume expansion generates more candidates that require increasingly costly verification.

\subsection{Human Difficulty Alignment with Model-Specific Gradients}
\label{sec:human_alignment}

The cross-family performance gaps and budget sensitivity patterns documented above raise a complementary question: do these differences track human difficulty perception? Using ground-truth data from 10,000 NYT Spelling Bee solvers per puzzle, we find that calibration follows the same cross-family hierarchy.

\paragraph{Calibration Strength Across Model Families}

We measure calibration using Spearman rank correlation between human difficulty and model miss rate (Figure~\ref{fig:human_calibration}). Among the four dense Qwen models, calibration is stable across scale ($\rho$ = 0.28 to 0.30, all $p < 0.001$), indicating that parameter scaling alone does not improve human difficulty alignment. The MoE 30B variant achieves notably higher calibration ($\rho$ = 0.36) at its F1-maximizing budget, consistent with its distinctive budget sensitivity profile (Section~\ref{sec:budget_sensitivity}). Proprietary models reach the highest alignment (Claude-Haiku $\rho$ = 0.42, GPT-5-mini $\rho$ = 0.41; all $p < 0.001$). The gap between dense Qwen models and proprietary models mirrors the cross-family F1 pattern from Section~\ref{sec:arch_scale}: the factors driving performance gains also improve human-like difficulty perception.

\paragraph{Difficulty Gradients and Uniform Performance Gaps}

Difficulty gradients vary sharply by capacity (Figure~\ref{fig:human_calibration}, left panel). Gradient steepness spans an order of magnitude across models, tracking the same capacity axis that drives recall differences (Section~\ref{sec:arch_scale}): models with limited recall lose a larger fraction of their retrieved set as difficulty increases. Scaling within the Qwen family improves robustness but does not eliminate the steep gradient, while the flatter proprietary gradients reflect the same broadly superior constraint-satisfaction capabilities that produce their recall advantage.

The proprietary models' absolute recall advantage persists uniformly across all difficulty quartiles, ruling out differential handling of difficult cases as the sole explanation. The uniform gap points to broadly superior constraint-satisfaction capabilities rather than merely better strategies for edge cases.

\begin{table}[t]
\centering
\small
\resizebox{\linewidth}{!}{
\begin{tabular}{l@{\hspace{0.4em}}c@{\hspace{0.4em}}c@{\hspace{0.4em}}c@{\hspace{0.4em}}l}
\toprule
\textbf{Word} & \textbf{Model Miss (\%)} & \textbf{Human Success (\%)} & \textbf{Puzzles} & \textbf{Pattern} \\
\midrule
loll       &  98.3 & 83.4 & 3 & Doubled cons. \\
illicit    &  98.3 & 84.3 & 3 & Doubled cons. \\
acai       &  97.4 & 84.2 & 3 & Loanword \\
nana       &  96.8 & 84.7 & 4 & Repeated letters \\
annotation &  96.6 & 86.1 & 3 & Doubled cons. \\
momma      &  96.6 & 87.6 & 3 & Doubled cons. \\
papa       &  95.7 & 88.8 & 3 & Repeated letters \\
ammo       &  94.0 & 89.5 & 3 & Doubled cons. \\
data       &  93.8 & 91.1 & 5 & Atypical pattern \\
toon       &  92.3 & 91.9 & 4 & Repeated letters \\
\bottomrule
\end{tabular}
}
\caption{Systematic failures: words with mean human success $>$80\% but high model miss rates ($>$92\%) across all configurations. Human success is averaged across puzzle appearances. Puzzles indicates how many of the 58 puzzles contained each word. Pattern categories: doubled consonants (ll, nn, mm), repeated letters (symmetric patterns), loanwords (borrowed terms with non-English orthographic origins), atypical patterns (common words whose letter patterns are underrepresented relative to their frequency).}
\label{tab:systematic_failures}
\end{table}

\paragraph{Systematic Failures and Two-Component Difficulty Structure}

Even the strongest correlations leave substantial unexplained variance. Words like ``data,'' ``toon,'' and ``papa'' (Table~\ref{tab:systematic_failures}) exemplify a recurring pattern: humans find these words easy yet models miss them at rates exceeding 92\%, despite clear vocabulary presence. These misses cannot be attributed to lexical gaps (all words appear frequently in pre-training corpora) but rather to failures in recognizing valid letter combinations under the specific constraint structure of each puzzle.

The moderate calibration coupled with substantial model-specific failures is consistent with partially overlapping difficulty drivers: (1) inherent linguistic difficulty that affects humans and models similarly, and (2) model-specific difficulty arising from capacity or training limitations. Proprietary models achieve both higher correlations and better absolute performance, indicating partial alignment between these components. Inter-model agreement on puzzle difficulty (mean $\rho = 0.457$; Appendix~\ref{sec:inter_model_agreement}) exceeds any individual model's alignment with human solvers, though this comparison spans different granularities (puzzle-level F1 vs.\ word-level miss rates) and the puzzle-level aggregation reduces noise. The direction nevertheless suggests that models share retrieval biases partially distinct from human difficulty drivers.

\subsection{Length-Dependent Performance and Systematic Difficulty Patterns}
\label{sec:length_patterns}

\begin{figure*}[t]
\centering
\includegraphics[width=0.96\textwidth]{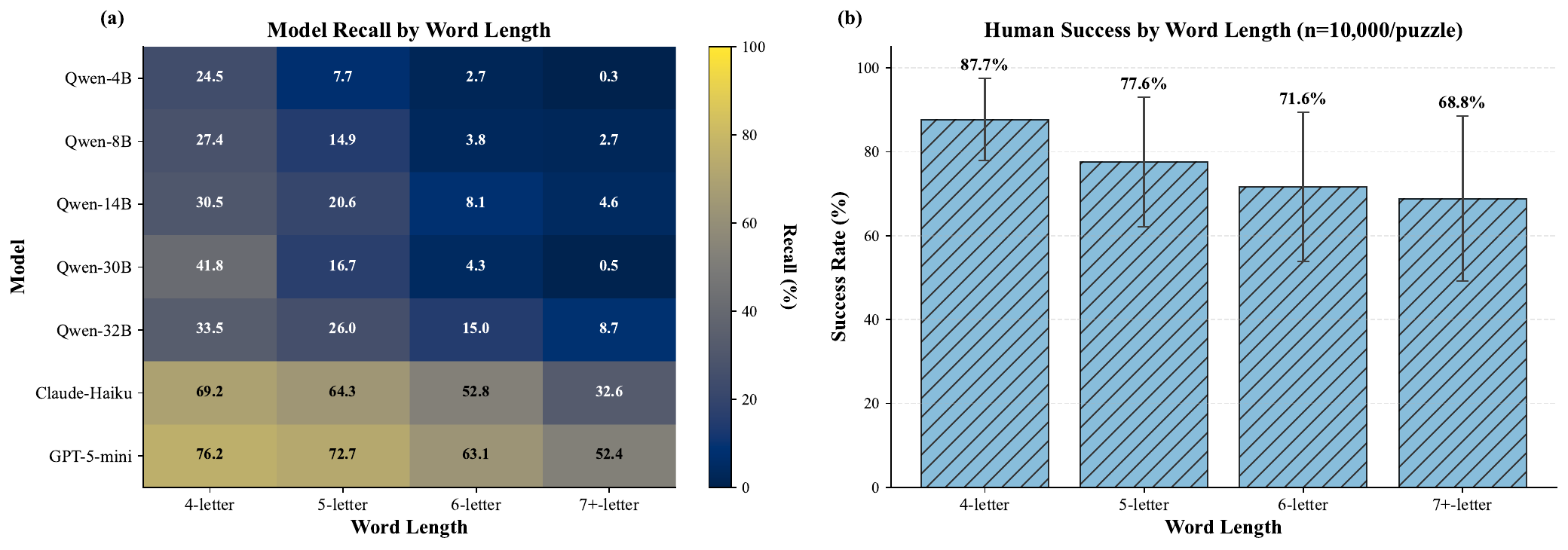}
\caption{Word length effects on model and human performance. Left: Model recall by word length. Right: Human success declines gently (1.3$\times$ drop) while models show catastrophic degradation (1.5--71$\times$+ drops).}
\label{fig:length_coverage}
\end{figure*}

The 60--80\% of variance unexplained by human difficulty (Section~\ref{sec:human_alignment}) points to model-specific failure modes. We investigate the structural properties behind this variance through two complementary lenses: categorical orthographic patterns that trigger systematic failures, and continuous length-dependent degradation that affects all models to varying degrees.

\paragraph{Categorical Orthographic Pattern Failures}

Table~\ref{tab:systematic_failures} identifies words that humans find easy yet get missed by models at extraordinarily high rates across all configurations. These failures cluster into three orthographic categories:

\begin{itemize}[nosep,leftmargin=*]
\item \textbf{Doubled consonants:} consecutive identical consonants such as ll, nn, mm (``illicit,'' ``loll,'' ``annotation,'' ``momma,'' ``ammo'').
\item \textbf{Repeated letters:} the same letter appearing multiple times non-consecutively, creating palindromic or symmetric patterns (``papa,'' ``nana,'' ``toon'').
\item \textbf{Loanwords and atypical patterns:} borrowed terms with non-English orthographic origins (``acai'') and common words whose letter patterns are underrepresented relative to their frequency (``data'').
\end{itemize}

The words in Table~\ref{tab:systematic_failures} appear frequently in training corpora, yet models miss them consistently across all configurations. Most are 4--5 letters long, the length range where models perform best overall. Orthographic atypicality overrides the length advantage, making these failures more diagnostic of the distributional plausibility bottleneck than the length gradient alone. These systematic retrieval failures persist across model families and scales.

\paragraph{Length-Dependent Performance Degradation}

Figure~\ref{fig:length_coverage} quantifies length-dependent patterns through direct model-human comparison. Performance stratified by word length reveals a consistent monotonic pattern: all models perform best on short words and progressively worse as length increases. This length gradient appears universal across model families but varies sharply in magnitude, with weaker models collapsing almost entirely on longer words while stronger models maintain substantial recall even at length extremes. Both humans and models exhibit monotonic decline as word length increases, confirming that length inherently increases difficulty. However, the magnitude of degradation differs dramatically: human success declines gently (1.3$\times$ from 4-letter to 7+ letter words), while model degradation ranges from 1.5$\times$ to over 71$\times$ depending on capacity. These architectures lack the robustness humans employ to maintain performance as combinatorial complexity grows. A partial-correlation analysis controlling for token count confirms that this degradation tracks character-level complexity rather than subword tokenization artifacts: within the Qwen family (shared tokenizer), token count adds no predictive power beyond character length (partial $r = -0.004$, $p = 0.845$; Appendix~\ref{sec:tokenization_check}).

\paragraph{Candidate Mechanistic Bottlenecks: Distributional Plausibility and Working Memory}

The observed failure patterns motivate two candidate computational bottlenecks. We present these as hypotheses that account for the empirical evidence rather than demonstrated mechanisms.

\textbf{Hypothesis 1: Over-reliance on distributional plausibility.} The categorical pattern failures indicate that constraint satisfaction may engage different mechanisms than natural language generation. In free-form text, words like ``data'' and ``papa'' appear in predictable semantic contexts that prime their generation. In constraint satisfaction tasks, models must verify character-level properties independently of contextual cues. The near-universal failures on doubled consonants (consecutive ll in ``illicit,'' nn in ``annotation,'' mm in ``momma'') and symmetric repeated patterns (``papa'' with alternating p-a-p-a, ``nana'' with n-a-n-a) implicate heavy reliance on distributional plausibility. These orthographically atypical but constraint-valid patterns may trigger lower generation probabilities than more common letter sequences, even when both satisfy constraints equally. We hypothesize that common words like ``data'' get systematically missed because the repeated `a' creates a d-a-t-a pattern with low generation probability under the model's learned distribution.

\textbf{Hypothesis 2: Limited working memory for multi-constraint verification.} The monotonic length-dependent decline tracks increasing verification load: each character in a candidate word requires an independent set-membership check against the seven allowed letters, plus verification that the mandatory letter appears at least once. A 4-letter word requires 4 such checks; a 9-letter word requires 9. While each check is simple, maintaining constraint state across more positions taxes sequential attention and increases opportunities for verification errors. Humans maintain 69\% success on 7+ letter words while small models collapse to 0.3\%, suggesting that smaller neural architectures lack the working memory capacity or systematic enumeration strategies humans employ. Testing this hypothesis would require controlled experiments varying constraint complexity while holding vocabulary constant.

\textbf{Implications.} Length-dependent patterns and systematic orthographic difficulties collectively demonstrate that constraint satisfaction failures are non-random. They cluster around structural properties (increasing length, unusual orthography) tied to specific computational bottlenecks: limited working memory for tracking multiple simultaneous constraints, over-reliance on distributional plausibility that penalizes valid-but-unusual patterns, and difficulty with systematic enumeration strategies. These failure modes motivate several directions: explicit constraint-tracking mechanisms that verify character-level validity independently of semantic plausibility, training objectives that reward generating orthographically unusual but valid solutions, and prompting strategies that decompose long-word generation into systematic prefix enumeration (i.e., building candidate words by extending valid letter prefixes one character at a time).

\section{Conclusion}

We investigated how large language models handle hard orthographic constraints, using the NYT Spelling Bee as a controlled testbed with human-calibrated difficulty data from 10,000 solvers per puzzle. Systematic evaluation across 39 configurations shows that cross-family differences (2.0--2.2$\times$) substantially exceed within-family scaling gains (83\%), manifesting through recall rather than precision. Thinking budget effects prove heterogeneous: the 14B model degrades with increased allocation while the MoE 30B variant (3B active parameters) requires substantial budget to compensate for its limited effective capacity.

Models systematically miss common words with atypical orthography (``data'': 94\% model miss vs.\ 91\% human success), revealing systematic retrieval failures despite vocabulary knowledge. Calibration with 10,000 human solvers is modest ($\rho$ = 0.28--0.42), with substantial unexplained variance indicating that model and human difficulty are driven by partially distinct factors.

These findings motivate four research directions: (1) explicit verification modules operating independently of distributional priors, (2) training objectives rewarding orthographically unusual but valid solutions, (3) model-specific budget allocation policies accounting for heterogeneous sensitivity, and (4) evaluating base model variants to isolate the effects of alignment training on constraint satisfaction, as censorship and safety tuning can alter generation behavior in task-relevant ways~\cite{tuck-verma-2025-unmasking}. If the candidate bottleneck our analysis highlights (over-reliance on distributional plausibility when structural validity is required) generalizes, it could be relevant to other constrained generation domains where correct outputs may be distributionally atypical, such as code synthesis with syntactic constraints, structured data generation requiring schema conformance, or mathematical reasoning demanding logical validity in unlikely intermediate steps.

\section{Limitations}

The experimental paradigm focuses exclusively on English orthography, limiting cross-linguistic generalization, though whether the distributional plausibility hypothesis transfers to other alphabetic systems remains untested. The task emphasizes lexical retrieval and constraint satisfaction without requiring semantic understanding, isolating one component of constrained generation while leaving semantic constraint interactions unexplored. Zero-shot evaluation reveals intrinsic constraint-handling capabilities; few-shot prompting might mitigate some failures but would not address underlying architectural limitations.

We evaluate on a single task domain (the NYT Spelling Bee). While this task provides a controlled testbed that enables precise measurement of constraint satisfaction with human-calibrated difficulty, generalization to other constrained-generation domains (crossword puzzles, anagram generation, code synthesis with syntax constraints) remains an open question. We treat the Spelling Bee as a methodological foundation; extending to additional domains is a natural next step.

Finally, the mechanistic hypotheses proposed in Section~\ref{sec:length_patterns} (distributional plausibility bias and working memory limitations) are generated by, not tested by, our empirical findings. Our partial-correlation analysis (Appendix~\ref{sec:tokenization_check}) rules out tokenization as a within-family confound and identifies only a modest cross-family residual, but validating the remaining hypotheses would require reasoning trace analysis to test whether models attempt systematic enumeration, and probing experiments to measure internal representations of character-level constraints.

\section{Ethical Considerations}

This work evaluates models on a publicly available word generation task, raising minimal ethical concerns. The experimental instances are drawn from the New York Times Spelling Bee for research and evaluation purposes consistent with fair use principles. We do not redistribute puzzle content beyond what is necessary for reproducibility. The word lists reflect editorial curation choices and may encode vocabulary biases, though the task focuses on orthographic constraints rather than semantic content.

\section*{Acknowledgments}

Research partly supported by NSF grant 2244279, ARO grant W911NF-23-1-0191 and a US Department of Transportation grant for CyberCare. Verma is the founder of Everest Cyber Security and Analytics, Inc.

\section{References}\label{sec:reference}

\bibliographystyle{lrec2026-natbib}
\bibliography{lrec2026-bib}

\newpage

\appendix
\section{Supplementary Results}

\subsection{Pangram Recall and Error Taxonomy}
\label{sec:pangram_fp}

\begin{table}[t]
\centering
\small
\begin{tabular}{l r r r}
\toprule
\textbf{Model} & \textbf{Budget} & \textbf{Pangram} & \textbf{Non-Dict} \\
 & & \textbf{Recall (\%)} & \textbf{FPs/puzzle} \\
\midrule
Qwen-4B & 16K & 1.7 & 1.4 \\
Qwen-8B & 4K & 2.7 & 1.9 \\
Qwen-14B & 4K & 10.5 & 2.4 \\
Qwen-30B & 8K & 0.0 & 2.3 \\
Qwen-32B & 8K & 15.8 & 2.6 \\
\midrule
Claude-Haiku & 16K & 47.3 & 4.3 \\
GPT-5-mini & 16K & \textbf{77.6} & 3.8 \\
\bottomrule
\end{tabular}
\caption{Pangram recall and non-dictionary false positives (words absent from any standard dictionary) per puzzle at each model's F1-maximizing budget.}
\label{tab:pangram_fp}
\end{table}

This section examines pangram recall and false positive composition to determine whether the cross-family performance gap from the main text extends to the hardest constraint-satisfaction subtask and whether model errors reflect genuine generation failures or vocabulary breadth.

Table~\ref{tab:pangram_fp} reports pangram recall and non-dictionary false positive rates at each model's F1-maximizing budget. The cross-family pangram recall gap between GPT-5-mini and the best open-source configuration (Qwen-32B) reaches 4.9$\times$. This substantially exceeds the 2.0--2.2$\times$ F1 gap reported in Section~\ref{sec:arch_scale}; notably, Qwen-30B fails to recall any pangrams at its F1-maximizing budget. Pangrams require all seven letters simultaneously, imposing a stricter combinatorial constraint that amplifies the cross-family divide.

No model in any configuration generates words violating the letter-set constraint: zero violations occur across all 39 configurations, verified programmatically by checking each generated word against the puzzle letter set (validation code included in the repository). The precision ceiling is instead bounded by lexical false positives: words satisfying the orthographic constraints but absent from the NYT's curated answer list.

\paragraph{False Positive Taxonomy}

Not all false positives represent the same type of error. A model-generated word absent from the NYT answer set may be either (a) a valid English word excluded by NYT editorial curation or (b) a fabricated string with no dictionary entry. These error types have different implications: the former reflects vocabulary breadth exceeding the curated answer set, while the latter reflects genuine generation failures.

To distinguish these cases, we collected all words generated across 39 configurations and 58 puzzles that did not appear in puzzle answer sets. All generated words were lowercased before lookup; no lemmatization was applied, so inflected forms present in either corpus were matched directly. We validated each against a reference dictionary constructed from the union of the NLTK words corpus and WordNet lemma set (259,404 entries). This reference dictionary provides broad coverage of standard English vocabulary but does not include all domain-specific, archaic, or borrowed terms; consequently, the valid-word rates in Table~\ref{tab:fp_taxonomy} are conservative lower bounds.

\begin{table}[t]
\centering
\small
\resizebox{\linewidth}{!}{
\begin{tabular}{l r r r r}
\toprule
\textbf{Family} & \textbf{Unique} & \textbf{Total} & \multicolumn{2}{c}{\textbf{Real English (\%)}} \\
\cmidrule(lr){4-5}
 & \textbf{words} & \textbf{instances} & \textbf{of unique} & \textbf{of occurrences} \\
\midrule
Qwen$^\dagger$ & 3,325 & 4,998 & 13.3 & 26.5 \\
Claude-Haiku$^\dagger$ & 2,746 & 4,489 & 17.0 & 26.0 \\
GPT-5-mini & 500 & 834 & \textbf{64.6} & \textbf{70.7} \\
\midrule
All models & 6,083 & 10,321 & 14.4 & 29.8 \\
\bottomrule
\end{tabular}
}
\caption{False positive taxonomy by model family. Unique words: distinct word forms not in the puzzle answer set. Total instances: summed across all puzzles and configurations. Real English (\% of unique): percentage of distinct false-positive word forms that are valid English words, validated against a 259K-entry reference dictionary (NLTK words corpus + WordNet lemmas). Real English (\% of occurrences): same validation applied to the total count of false-positive instances across all puzzles and configurations; a higher rate here than in the unique column indicates that real-English false positives are generated more repeatedly than fabricated ones. Both rates are conservative lower bounds as the dictionary excludes some domain-specific and archaic terms. $^\dagger$Includes both direct and thinking mode configurations; GPT-5-mini lacks a direct mode baseline.}
\label{tab:fp_taxonomy}
\end{table}

Table~\ref{tab:fp_taxonomy} breaks down false positives by family. The gap between the two Real English columns (higher by occurrences than by unique words) confirms that valid-English false positives recur more frequently across configurations than fabricated ones, consistent with models drawing on common vocabulary items that happen to fall outside NYT's curated lists.

The cross-family divergence is sharp: GPT-5-mini produces the fewest unique FP types but the highest valid-word rate, dominated by low-frequency but legitimate English words (e.g., ``chine,'' ``milt,'' ``adit''). Qwen models produce far more unique FP types, the vast majority of which are fabricated strings. GPT-5-mini's profile reflects broader vocabulary coverage exceeding NYT's curated lists; Qwen's reflects genuine generation failures. Raw false positive counts therefore conflate two qualitatively different error types, and unadjusted precision metrics understate the vocabulary quality of models with broader lexical coverage.

\subsection{Generation Strategy Under Extended Reasoning}
\label{sec:strategy_profiles}

This section examines how thinking mode changes generation behavior, distinguishing models that expand output volume from those that filter it.

\begin{table}[t]
\centering
\small
\resizebox{\linewidth}{!}{
\begin{tabular}{l @{\hspace{0.6em}} c @{\hspace{0.6em}} r @{\hspace{0.6em}} r @{\hspace{0.6em}} r @{\hspace{0.6em}} r}
\toprule
\textbf{Model} & \textbf{Think} & \textbf{Vol} & \textbf{$\Delta$Vol} & \textbf{F1} & \textbf{$\Delta$F1} \\
\midrule
\multicolumn{6}{l}{\textbf{Qwen3 Family}} \\
\addlinespace[2pt]
4B  & OFF & 2.9 &       & 0.048 &  \\
    & ON  & 6.5 & +122\% & 0.187 & +0.139 \\
\addlinespace[2pt]
8B  & OFF & 5.9 &       & 0.059 &  \\
    & ON  & 8.4 & +43\%  & 0.235 & +0.177 \\
\addlinespace[2pt]
14B & OFF & 9.6 &       & 0.060 &  \\
    & ON  & 10.6 & +10\% & 0.289 & +0.229 \\
\addlinespace[2pt]
30B & OFF & 4.0 &       & 0.072 &  \\
    & ON  & 11.2 & +178\% & 0.310 & +0.238 \\
\addlinespace[2pt]
32B & OFF & 5.8 &       & 0.116 &  \\
    & ON  & 12.7 & +118\% & 0.343 & +0.227 \\
\midrule
\multicolumn{6}{l}{\textbf{Proprietary Models}} \\
\addlinespace[2pt]
Claude-Haiku & OFF & 37.5 &      & 0.360 &  \\
             & ON  & 30.4 & $-$19\% & 0.680 & +0.321 \\
\addlinespace[2pt]
GPT-5-mini   & ON  & 34.4 &      & 0.761 &  \\
\bottomrule
\end{tabular}
}
\caption{Generation strategy profiles by thinking mode at each model's F1-maximizing budget. Vol: mean words generated per puzzle. Think: thinking mode ON or OFF. $\Delta$Vol and $\Delta$F1 show percentage and absolute change from OFF to ON, computed as (ON $-$ OFF) / OFF $\times$ 100 for $\Delta$Vol and ON $-$ OFF for $\Delta$F1. GPT-5-mini lacks an OFF baseline because all tested configurations use reasoning.}
\label{tab:strategy_profiles}
\end{table}

Table~\ref{tab:strategy_profiles} reports generation volume and F1 under thinking ON versus OFF at each model's F1-maximizing budget. Two distinct strategies emerge.

\paragraph{Volume-Increasing Pattern (Qwen Family)}
All five Qwen models increase output volume under thinking, generating more candidate words and discovering more valid solutions but also introducing additional false positives. Volume increase magnitude does not predict F1 gain (Table~\ref{tab:strategy_profiles}): the model that expands most does not achieve the largest F1 improvement, while the model that expands least achieves a comparable gain. Volume expansion is therefore necessary but not sufficient; the precision of expanded candidates varies by model capacity.

\paragraph{Volume-Decreasing Pattern (Claude-Haiku)}
Claude-Haiku reverses the pattern: volume decreases under thinking while F1 nearly doubles. This filtering behavior suppresses invalid candidates rather than generating additional ones. Proprietary models generate higher absolute volume than Qwen models, producing more total false positives per puzzle (Table~\ref{tab:pangram_fp}), but this reflects generation volume rather than lower precision (Table~\ref{tab:cross_family}). The false positive taxonomy in Section~\ref{sec:pangram_fp} confirms that most GPT-5-mini false positives are valid English words rather than fabrications.

\paragraph{Connection to Budget Sensitivity}
These patterns explain the budget sensitivity reported in Section~\ref{sec:budget_sensitivity}. Volume expansion produces diminishing returns because additional tokens generate more candidates without proportionally improving constraint verification, consistent with the plateau and degradation observed in Qwen models at higher budgets. Volume-decreasing behavior yields consistent gains: additional reasoning tokens appear to remove false positives from the candidate set, consistent with Claude-Haiku's monotonic improvement across budgets.

\paragraph{Complete Generation Collapse}
Without thinking, Qwen-4B produces zero valid words on 50.6\% of puzzles (88 of 174 puzzle-configuration pairs across three budgets), compared to 0\% for all proprietary configurations. Failure rates vary non-monotonically by budget (46.6\% at 4K, 60.3\% at 8K, 44.8\% at 16K), with the highest failure rate at the intermediate budget rather than the smallest. This pattern suggests that output length variation interacts with generation quality in model-specific ways and cannot rescue a model below the capacity threshold for constraint satisfaction. This complete generation collapse, rather than gradual degradation, produces the bimodal F1 distributions analyzed in Section~\ref{sec:f1_distributions}.

\subsection{Performance Distribution Characteristics}
\label{sec:f1_distributions}

Mean F1 scores can obscure qualitative reliability differences between models. This section examines per-puzzle F1 distributions to identify failure modes hidden by averages.

\begin{figure*}[t]
\centering
\includegraphics[width=0.92\textwidth]{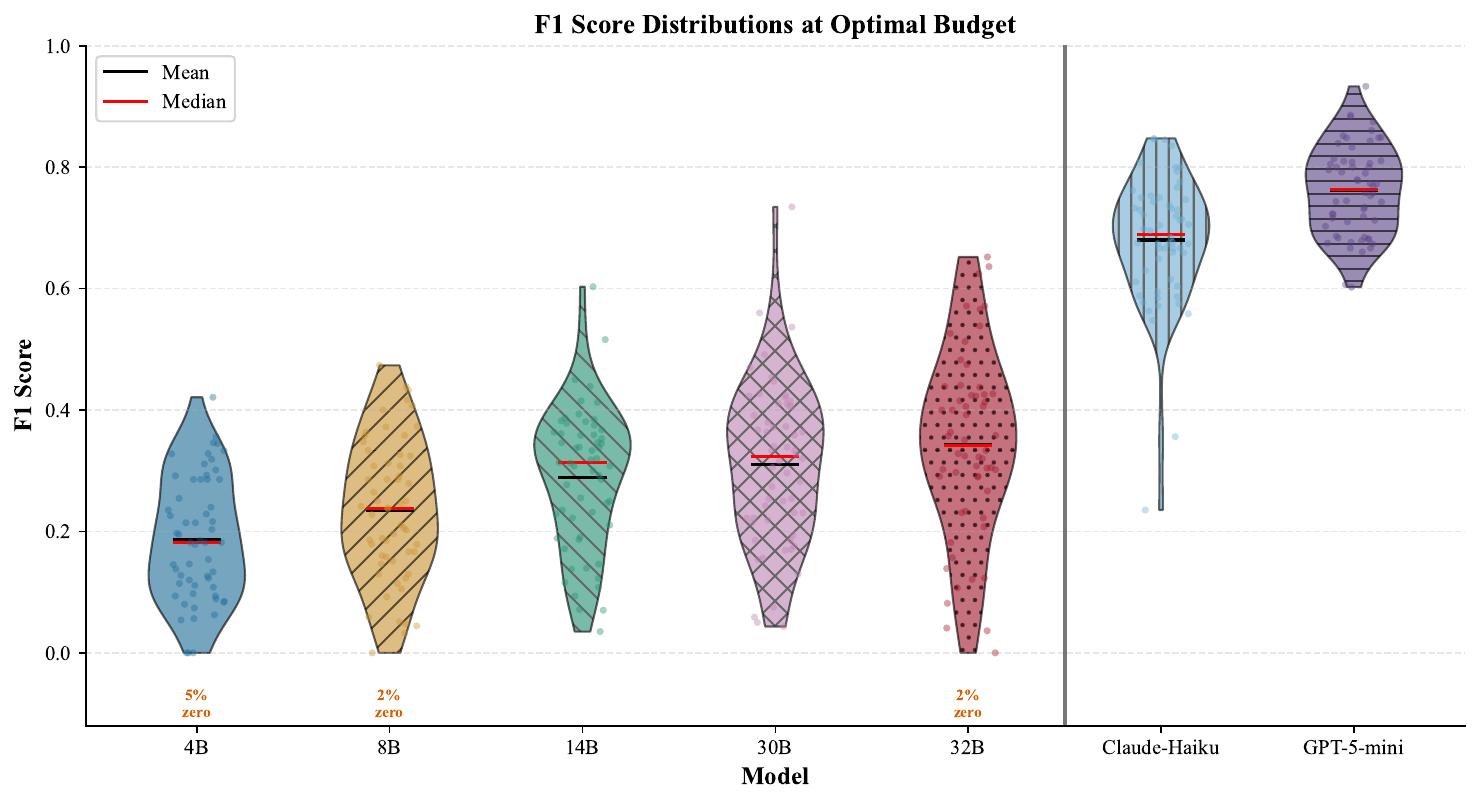}
\caption{Per-puzzle F1 score distributions at each model's F1-maximizing budget. Violin width encodes density; black lines show means, red lines show medians. Percentages below violins indicate complete failure rates ($F_1 = 0$).}
\label{fig:f1_distributions}
\end{figure*}

Figure~\ref{fig:f1_distributions} reveals reliability differences that mean F1 scores obscure. Proprietary models achieve high means through consistency: tight distributions concentrated near the ceiling, with zero complete failures ($F_1 = 0$) across all 58 puzzles. Qwen models produce qualitatively different distributions, exhibiting mixtures of moderate successes and complete failures rather than uniform moderate performance. Even at best configurations, open-source models occasionally produce zero valid words, a failure mode entirely absent from proprietary results.

These failure rates are low in absolute terms but represent a categorically different reliability profile. For deployment, this distinction carries practical weight: a system that occasionally produces zero valid outputs requires different error-handling than one that consistently produces partial solutions, even at identical mean F1.

\subsection{Inter-Model Agreement on Puzzle Difficulty}
\label{sec:inter_model_agreement}

This section asks whether models agree on which puzzles are difficult, and how that agreement compares to model-human alignment.

\begin{figure}[t]
\centering
\includegraphics[width=\linewidth]{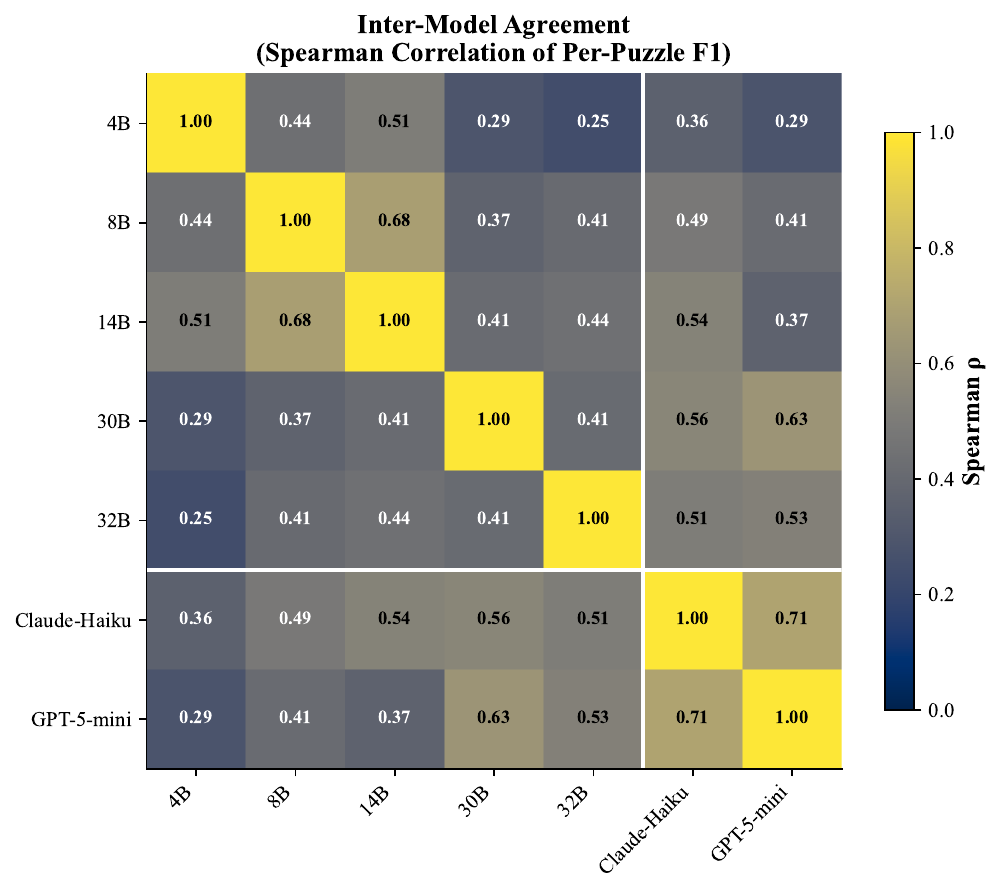}
\caption{Pairwise Spearman rank correlation of per-puzzle F1 scores across models at each model's F1-maximizing budget. All pairwise correlations significant at $p < 0.001$. Mean pairwise $\rho = 0.457$. White lines separate Qwen (open-source) from proprietary model families.}
\label{fig:inter_model_agreement}
\end{figure}

Figure~\ref{fig:inter_model_agreement} shows pairwise Spearman rank correlations of per-puzzle F1 scores across all seven models at each model's F1-maximizing budget ($n = 58$ puzzles per pair). The mean pairwise $\rho = 0.457$ exceeds any individual model's correlation with human difficulty ($\rho$ = 0.28--0.42, Section~\ref{sec:human_alignment}). However, the two comparisons operate at different granularities: inter-model agreement is measured at the puzzle level, while human-model calibration is measured at the word level. Puzzle-level aggregation reduces noise and inflates correlations, so the difference in magnitude should not be over-interpreted. The direction nevertheless suggests that models share puzzle-level difficulty drivers that partially diverge from the factors affecting human solvers. Plausible sources include common training data distributions, shared tokenizer-induced biases, or similar failure modes on orthographically atypical letter sets.

Agreement structure varies systematically across the heatmap. Within the Qwen family, small-to-large correlations are modest, indicating that scaling changes which puzzles a model finds difficult rather than uniformly improving all puzzles. Claude-Haiku and GPT-5-mini show the strongest pairwise agreement, consistent with their similar absolute performance levels and filtering-oriented generation strategies. The cross-family block structure (higher within-family than between-family correlations for Qwen) suggests that architecture-specific factors influence which puzzles are difficult, in addition to factors shared across all models.

\subsection{Tokenization Robustness Check}
\label{sec:tokenization_check}

Section~\ref{sec:length_patterns} reports that model performance degrades monotonically with word length. Because subword tokenizers encode longer words into more tokens, observed length-dependent degradation could reflect tokenization granularity (more tokens = more decoding steps = more opportunities for error) rather than character-level combinatorial complexity. To distinguish these accounts, we tokenized all 2,007 unique target words using the Qwen3 tokenizer (shared across all five Qwen models) and analyzed the independent contributions of character length and token count to model success rates.

\paragraph{Token Count Distribution}

Tables~\ref{tab:token_dist} and~\ref{tab:tokens_by_length} report the distribution of target words by token count and character length, with recall averaged across all 39 configurations. The majority of target words require multiple subword tokens, and mean token count increases monotonically with character length. The Pearson correlation between the two predictors is moderate ($r = 0.466$, $n = 2{,}007$): high enough that shared variance must be accounted for, but far from collinear, making partial correlation a viable method to isolate each predictor's independent contribution.

\begin{table}[t]
\centering
\resizebox{\linewidth}{!}{
\begin{tabular}{cccc}
\toprule
\textbf{Token count} & \textbf{Words} & \textbf{\%} & \textbf{Recall} \\
\midrule
1 & 429 & 21.4 & 0.250 \\
2 & 1,281 & 63.8 & 0.179 \\
3 & 286 & 14.3 & 0.072 \\
4 & 11 & 0.5 & 0.056 \\
\bottomrule
\end{tabular}
}
\caption{Distribution of Spelling Bee target words by Qwen3 subword token count, with aggregate recall across all model configurations.}
\label{tab:token_dist}
\end{table}

\begin{table}[t]
\centering
\resizebox{\linewidth}{!}{
\begin{tabular}{ccccc}
\toprule
\textbf{Char length} & \textbf{$n$} & \textbf{Mean tokens} & \textbf{Single-token \%} & \textbf{Recall} \\
\midrule
4 & 589 & 1.58 & 42.4 & 0.275 \\
5 & 447 & 1.86 & 19.9 & 0.210 \\
6 & 402 & 2.05 & 11.7 & 0.120 \\
7 & 279 & 2.16 & 8.2 & 0.087 \\
8 & 157 & 2.36 & 7.0 & 0.056 \\
9 & 69 & 2.41 & 8.7 & 0.048 \\
10+ & 64 & 2.59 & 4.7 & 0.030 \\
\bottomrule
\end{tabular}
}
\caption{Tokenization profile and aggregate recall by character length. $n$: number of unique words at each length. Recall drops monotonically with character length; single-token percentage also drops, though the two gradients are only moderately correlated.}
\label{tab:tokens_by_length}
\end{table}

\paragraph{Partial Correlation Analysis}

To isolate each metric's independent contribution, we computed partial correlations of word-level success rate with character length and token count, each controlling for the other. Success rates are averaged across configurations for each unique word ($n = 2{,}007$), treating each word as a single observation. We report results at two scopes (Table~\ref{tab:tokenization}): all models combined (where cross-family tokenizer differences may contribute) and the Qwen family alone (shared tokenizer, isolating the within-family confound question).

\begin{table}[t]
\centering
\resizebox{\linewidth}{!}{
\begin{tabular}{llcc}
\toprule
\textbf{Scope} & \textbf{Predictor (controlling for)} & \textbf{Partial $r$} & \textbf{$p$} \\
\midrule
\multirow{2}{*}{All models (39 configs)} & Char length $\mid$ token count & $-0.446$ & $<10^{-98}$ \\
& Token count $\mid$ char length & $-0.110$ & $<10^{-6}$ \\
\midrule
\multirow{2}{*}{Qwen only (30 configs)} & Char length $\mid$ token count & $-0.399$ & $<10^{-77}$ \\
& Token count $\mid$ char length & $-0.004$ & $0.845$ \\
\bottomrule
\end{tabular}
}
\caption{Partial correlations of word-level success rate with character length and token count (2,007 unique words). Within the Qwen family (shared tokenizer), token count has no residual predictive power after controlling for character length.}
\label{tab:tokenization}
\end{table}

The Qwen-only analysis provides the cleanest test because all five models share an identical tokenizer, eliminating cross-tokenizer confounds. As Table~\ref{tab:tokenization} shows, token count has no residual predictive power beyond character length for within-family comparisons, while character length retains a robust independent effect at both scopes. This confirms that character-level combinatorial complexity, not subword segmentation, drives the observed length-dependent degradation.

\begin{table}[htbp]
\centering
\resizebox{\linewidth}{!}{
\begin{tabular}{ccccc}
\toprule
\textbf{Char length} & \textbf{$r$ (token count)} & \textbf{$p$} & \textbf{1-token recall} & \textbf{2-token recall} \\
\midrule
4 & $-0.038$ & $0.362$ & 0.307 & 0.297 \\
5 & $-0.139$ & $0.003$ & 0.271 & 0.219 \\
6 & $-0.213$ & $<0.001$ & 0.150 & 0.137 \\
7 & $-0.084$ & $0.164$ & 0.126 & 0.091 \\
\bottomrule
\end{tabular}
}
\caption{Within-length-stratum Pearson correlation of token count with word-level success rate. Significant effects appear only at 5--6 letters; 4-letter and 7-letter words show no significant token count effect.}
\label{tab:within_length}
\end{table}

\paragraph{Within-Length Strata}

As a further check, we examined whether token count predicts success within words of the same character length, removing between-length variance entirely (Table~\ref{tab:within_length}). Significant token-count effects appear only at intermediate lengths (5--6 letters), where both single- and multi-token encodings are common enough to support comparison. At shorter and longer lengths, the effect is non-significant. The absolute recall differences between 1-token and 2-token words within each stratum are small (1--5 percentage points), compared to the 25-percentage-point recall drop between 4-letter and 7-letter words overall. Character-level complexity remains the primary driver; tokenization contributes a secondary effect only where encoding variation is sufficient to detect it.

Across all analyses, character length consistently dominates token count as a predictor of model difficulty. The cross-family residual (Table~\ref{tab:tokenization}) indicates that tokenizer differences contribute modestly to performance gaps between families but do not account for the core length-dependent degradation pattern.

\end{document}